\documentclass[conference]{IEEEtran}

\IEEEoverridecommandlockouts
\usepackage{arabtex}
\usepackage{utf8}
\setcode{utf8}

\usepackage{cite}
\usepackage{amsmath,amssymb,amsfonts}
\usepackage{algorithmic}
\usepackage{graphicx}
\usepackage{textcomp}
\usepackage{tablefootnote}

\usepackage{fancyhdr}

\usepackage[utf8]{inputenc} 
\usepackage{hyperref}       
\usepackage{url}            
\usepackage{booktabs}       
\usepackage{amsfonts}       
\usepackage{nicefrac}       
\usepackage{microtype}      
\usepackage{lipsum}		
\usepackage{graphicx}
\usepackage{doi}
\usepackage{xcolor}
\usepackage{float}
\usepackage{pdfpages}

\usepackage{multirow}
\usepackage{booktabs, multirow} 
\usepackage{soul}
\usepackage{atbegshi} 
\AtBeginDocument{\AtBeginShipoutNext{\AtBeginShipoutDiscard}}

\makeatletter

\def\ps@IEEEtitlepagestyle{
  \def\@oddfoot{\mycopyrightnotice}
  \def\@evenfoot{}
}
\def\mycopyrightnotice{
  {\footnotesize \hfill} 
  \gdef\mycopyrightnotice{}
}

\newcommand*\titleheader[1]{\gdef\@titleheader{#1}}
\AtBeginDocument{%
  \let\st@red@title\@title
  \def\@title{%
    \bgroup\normalfont\large\centering\@titleheader\par\egroup
    \vskip1.5em\st@red@title}
}

\title{Word-level Persian Lipreading Dataset}
\titleheader{}

\graphicspath{ {./} }

\def\BibTeX{{\rm B\kern-.05em{\sc i\kern-.025em b}\kern-.08em
    T\kern-.1667em\lower.7ex\hbox{E}\kern-.125emX}}
\begin{document}.

\author{
\IEEEauthorblockN{Javad Peymanfard\,\IEEEauthorrefmark{1},
Ali Lashini\,\IEEEauthorrefmark{1}\textsuperscript{\textsection},
Samin Heydarian\,\IEEEauthorrefmark{1}\textsuperscript{\textsection}, 
Hossein Zeinali\,\IEEEauthorrefmark{2}, and
Nasser Mozayani\,\IEEEauthorrefmark{1}}
\vspace{2mm}
\IEEEauthorblockA{\IEEEauthorrefmark{1}\,School of Computer Engineering\\
Iran University of Science and Technology, Tehran, Iran \\
Email: \texttt{javad\_peymanfard@comp.iust.ac.ir, a\_lashini@comp.iust.ac.ir}\\
\texttt{samin\_heydarian@comp.iust.ac.ir, mozayani@iust.ac.ir}
}
\vspace{2mm}
\IEEEauthorblockA{\IEEEauthorrefmark{2}\,Department of Computer Engineering \\
Amirkabir University of Technology, Tehran, Iran \\
Email: \texttt{hzeinali@aut.ac.ir}
}
}

\maketitle

\maketitle

\begingroup\renewcommand\thefootnote{\textsection}
\footnotetext{These authors contributed equally.}
\endgroup

\begin{abstract}
Lip-reading has made impressive progress in recent years, driven by advances in deep learning. Nonetheless, the prerequisite such advances is a suitable dataset. This paper provides a new in-the-wild dataset for Persian word-level lip-reading containing 244,000 videos from approximately 1,800 speakers. We evaluated the state-of-the-art method in this field and used a novel approach for word-level lip-reading. In this method, we used the AV-HuBERT model for feature extraction and obtained significantly better performance on our dataset.
\end{abstract}


\begin{figure*}[t]
  \centering
  \includegraphics[width=0.9\textwidth]{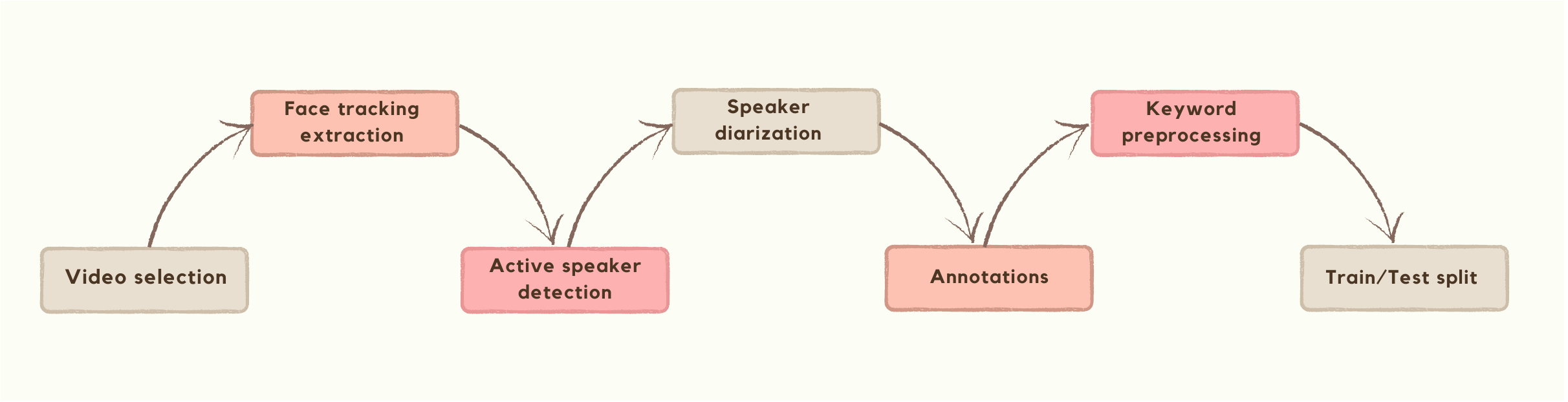}
  \caption{Data collection pipline.}
  \label{fig:data collection pipline}
\end{figure*}

\section{Introduction}
Lip-reading, also known in the research community as visual speech recognition (VSR), is the task of recognizing words or characters being said from a sequence of lip movements without any audio streams. It has proven to be quite useful in situations such as when audio signals are corrupted by noise. In these cases, VSR can be incorporated with speech recognition models in order to improve accuracy.

The traditional approaches consist of two stages. The first stage is called backbone (or front-end) and the second stage is called back-end. In front-end, a sequence or combination of images are extracted and then passed to another network to produce words or sentences.~\cite{stafylakis2017combining, chung2017lip, peymanfard2022lip} The procedure done for back-end is very similar to the procedure in Automatic Speech Recognition (ASR) methods. However, because of the existence of more large-scale datasets and ease of task compared to lip-reading, there is a significant gap in performance between the two models.

Primary datasets were collected under supervised conditions (laboratory).~\cite{5745028, matthews2002extraction, 4a49fd04cf0b4184b9f780209e099de1} These particular datasets were usually small and were limited to simple tasks such as character and digit classification. Although, with further advances in deep learning, these primary datasets no longer provided a challenge. Researchers began to focus on building large-scale and more challenging datasets; no longer collected under supervised conditions. They were collected from TV shows, movies, interviews, etc. In other words, there was no longer control over the condition of the speakers and the surrounding environment, creating what is called the "wild" condition. Recently collected datasets can be fit into one of the following categories: 1) word-level datasets, and 2) sentence-level datasets.~\cite{chung2017lip, chung2017lipmv, afouras2018lrs3} In this paper, we will focus on lip-reading in the word-level stage. LRW~\cite{chung2016lip} and LRW-1000~\cite{yang2019lrw} in English and Chinese language respectively are two popular datasets in this field.

Currently, LRW-1000 is the largest dataset in this field, as well as the most challenging. It contains 1000 classes with 718,018 samples of Mandarin words from over 2000 speakers, with videos that are not fixed in resolution. The highest accuracy achieved without extra data on LRW-1000 was 53.8 percent by (Kim. M, etc.)~\cite{kim2022distinguishing}. 
On the other hand, LRW is the most widely used dataset that contains 500 classes with about 800 samples per class collected from BBC TV programs. The highest accuracy achieved here was also by (Kim. M, etc.)~\cite{kim2022distinguishing} with 88.5 percent. GLips~\cite{schwiebert2022multimodal} is the most recent dataset in this field released in the German language and is very similar to LRW. This dataset contains 500 classes with 500 samples per class collected from YouTube.

In this paper, we want to propose a novel word-level dataset for lip-reading in the Persian language. To collect our data, we used a Persian video streaming website called Aparat. The dataset consists of various light conditions, poses, etc. The videos include various types of TV and internet shows such as interviews, movies, etc., resulting in 205 hours of videos. We provide 500 of the most frequent words as labels to our dataset. The samples in the dataset are various in length and characters. Due to ambiguity in the Persian language and from the difficulty proven from the experiment that we have done on the recent method; this dataset could be as challenging, if not more, as LRW-1000. The experiments with the recent methods and different conditions and metrics are available in the experiment section in part four of this paper.

In the continuation of this paper, we intend to deal with the following sections: First, we will look at some of the previous works in this field, and after that, we are going to describe the statistics of data and the pipeline we used to collect data. To conclude, we will show some baseline results on the dataset.

\begin{figure*}[t]
  \centering
  \includegraphics[width=0.9\textwidth]{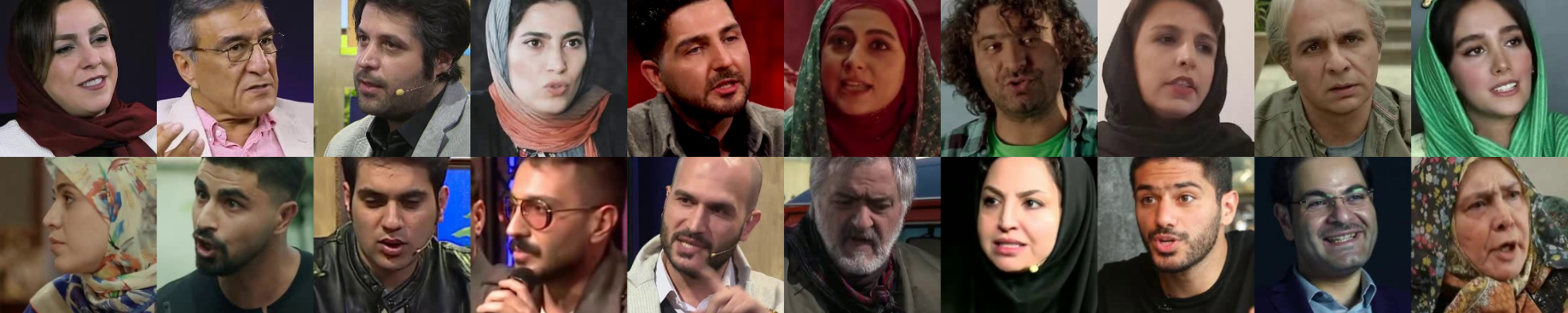}
  \caption{Our dataset samples.}
  \label{fig:dataset_samples}
\end{figure*}

\section{Related Works}
In this section, we are going to examine some of the prominent approaches in the field of word-level lip-reading followed by an overview of some of the datasets that have recently been used for word-level lip-reading tasks.

In recent approaches to building a network, usually the number of 3D and 2D convolutional networks are used as the front-end to extract visual feature. In the back-end part of the procedure, usually recurrent neural networks, transformers, etc., are used to predict words. One of the most prominent works is done by T. Stafylakis and G. Tzimiropoulos~\cite{stafylakis2017combining}. They used a combination of 3D convolution and ResNet with 34 layers as front-end and Bi-directional LSTM as back-end. The method became state-of-the-art in 2017 and the front-end network was used as a backbone for many future works after that. MS-TCN~\cite{martinez2020lipreading} is another network that uses a 3D convolution and ResNet with 18 layers as the front-end and a Temporal convolution network as the back-end. Recently, knowledge distillation has been used in this field~\cite{ma2021towards,kim2022distinguishing}, and MS-TCN is adopted as the main network for their work.

\textit{LRW}~\cite{chung2016lip} is a large-scale dataset which is the most widely used in the field of word-level lip-reading, released in 2016. The dataset contains 500 classes, and each class has at least 800 samples. The videos were collected from a UK BBC TV news program with more than a thousand speakers, bringing about one of the first datasets collected under the “wild” condition. Each video has a fixed length of 1 minute and 16 seconds, and the target class is in the center of the video. The dataset is quite challenging because some classes are very similar to each other, such as SPENT and SPEND (both classes have the same visemes), making it difficult for the classifier to distinguish these two words. This dataset is the largest dataset in the English language.

\textit{LRW-1000}~\cite{yang2019lrw} is another large-scale dataset which was collected from 51 programs broadcasted in China. The dataset contains 718,018 videos with 508 hours of videos in total. It also includes 1000 classes with 718 samples per class. The minimum duration of each word is 0.01 seconds, and the maximum duration is 2.25 seconds. The classes are commonly used Chinese words. The resolution of the video is not fixed and includes a variety of resolutions from standard-definition of 1024*576 to high-definition of 1920*1080 captured in 25 fps. The dataset contains a large variety of different conditions, including variations in age, posture, etc. This is not only the largest dataset in the Chinese language, but also the largest in the word-level lip-reading field as well as the most challenging because of the points mentioned above. 

\textit{GLips}~\cite{schwiebert2022multimodal} is a novel large-scale dataset that was collected in the German language. This dataset is very similar to LRW. The dataset contains 250 thousand videos collected from YouTube, each video having a length of 1.16 seconds long and a resolution of 256*256, as it is in LRW. They used 500 target words as labels with lengths of 4-18 characters, and each word has 500 instances. Besides the target word, they provided a start, end, and the duration of utterance in seconds as labels too. Also, the audio is separated from the video to be used for different purposes. This is the largest lip-reading dataset in the German language.

\textit{OuluVS}~\cite{zhao2009lipreading} is slightly different from the datasets mentioned above. This dataset was released in 2009 and contained phrases instead of words, and is used for phrase classification. The dataset was collected in a “laboratory” condition and contained ten classes (phrases) with 817 samples in total spoken by 20 persons. For the convenience of researchers and to avoid additional preprocessing steps, the authors provide a cropped region of the mouth for videos in 80*60 resolution. The dataset was widely used in its time, but presently, because of advances in deep learning methods and the small size of the dataset, it has become less challenging for the research community. 

\textit{OuluVS2}~\cite{anina2015ouluvs2} was released six years after OuluVS in 2015. In this dataset, the number of classes stays the same as the previous one, but the number of speakers increased to 53 persons and 1060 samples in total. To make it more challenging, they increased the resolution from 80*60 to 1920*1080. Additionally, the videos were shot from 5 different angles (0, 30, 45, 60, and 90 degrees). Similar to the previous dataset, this one is not commonly used nowadays. 

\section{Data Collection}

\subsection{Data Description}


The dataset includes over 500 vocabularies and 30 hours of clips of more than 1800 celebrities that were collected from videos posted to the Persian website Aparat. Videos included in the training and validation sets contain 233K samples, and the test set contains 11K samples. Also, The videos are captured in a variety of demanding audio and visual surroundings. The videos are provided as mp4 files and consist of cropped faces with a resolution of $224\times 224$ and a frame rate of 25 fps. Table \ref{tab:parsbertevaluation} shows the general statistics, and figure \ref{fig:dataset_samples} shows some examples of the video samples.


\begin{table*}
 \caption{Statistical comparison between different datasets}
  \centering
  \setlength\tabcolsep{12pt}
  \begin{tabular}{l | l | l | l | l}
    \toprule
    \toprule
    Dataset     & PLRW     & LRW    &  LRW-1000 & GLips \\
    \midrule
        Classes & 500 & 500 & 1000 & 500   \\
        Word length & 3-8 & 5-12 & - &  4-18     \\
        \# of speakers & $\>1800$ & $\>1000$ & $\>2000$ & $\sim100$  \\
        \# of videos & $\sim244,000$ & $\sim500,000$ & 718,018 & $\sim250,000$  \\
        Environment & Wild & Wild & Wild & Wild  \\
        Resolution & 224*224 & 256*256 & Various & 256*256 \\
        Video Length & 0.01-2.6s & 1,16s & 0.01-2.25s & 1,16 \\
        Frame rate & 25fps & 25fps & 25fps & 25fps \\
        Language & Persian & English & Chinese & German \\
        Time & 30h & 173h & 508h & - \\
        Source & Aparat & BBC TV & TV Shows & YouTube \\
        Year & 2022 & 2016 & 2018 & 2022 \\
    \bottomrule
    \bottomrule
  \end{tabular}
  \label{tab:parsbertevaluation}
\end{table*}

\subsection{Automatic Pipeline}
\textbf{Stage 1: Video selection.}
    For our dataset, we choose one of three types of videos, as follows. 
    \begin{enumerate}
      \item \textbf{Interview videos.}
      For audio-visual datasets, interviews and biographies work well. The major objective of these videos is to interview renowned people, whether they be actors, musicians, or politicians. There are two categories for Interview videos. The shows in the first category have hosts and narrators (which is not ideal in this circumstance), whereas those in the second category do not. Thus, the narrator should be removed from the first set of videos because our goal is have an audio-visual dataset. The host should be pruned too. There are also some breaks in the second category that need to be identified and deleted. 
      \item \textbf{Movies and series.}
      Movies and series can be utilised if appropriate preprocessing is done to them. The key difficulty is camera perspectives (Unlike in interviews, a speaker is not always in the shot while speaking). In addition, the entire video contains a considerable quantity of silence or music. In addition, Multi-speaker simultaneous speech presents another difficulty. Due to their high level of duplicated data, this form of video is often not the best for data collection. We found that Less than 10\% of the input data is deemed suitable.
      \item \textbf{Vod videos.}
      Another method for data collection is to use various keywords to search in ugcs, analyze the videos that come up, and then select the best keywords.
    \end{enumerate}
\textbf{Stage 2: Face tracking extraction.}
At this point, the video's various scenes are identified by "The difference between the pixel values of two successive frames." These differences were computed using the HSV color space. The threshold for a new scene is determined by the heuristic.       
As a result, each video is broken into smaller segments known as scenes. In fact, a temporal split was applied to the video so far. In the following step, the frame which contains the face is important. We utilized the described method in ~\cite{zhang2017s3fd} for face detection and an IoU based algorithm~\cite{tao2021someone} for face tracking, so that some frames of each scene are chosen. Each video may be divided into one or more smaller videos based on the number of faces detected in each segment, or it may be eliminated if there are none.

\textbf{Stage 3: Active speaker detection.}
    One of the most crucial analytical steps for producing the appropriate dataset is this stage. Active speaker detection determines which (if any) of the observable individuals in a video are speaking at any given moment. So a speaker is called active when his or her face is visible and the voice is audible at the same time. For this, we use TalkNet~\cite{tao2021someone}. This model was trained on the AVA dataset~\cite{roth2020ava}, which is roughly forty hours with precise labeling and is composed of a variety of languages. They utilize YouTube videos as the source for collecting the dataset. The TalkNet model detects active speakers on Persian Videos quite effectively, according to tests conducted on a sizable Persian dataset.
    
\textbf{Stage 4: Multi-speaker detection based on audio analysis.}
     The video segments with a fixed recording angle and a speaking subject have been selected till Stage 4. Speaking of other non-active speakers at the same time as the active speaker presents another possible difficulty. In interviews, it more frequently occurs when the host speaks while the visitor is speaking. Hence Speaker diarization can help solve this issue. The input audio is clustered by this technique and each cluster carries a speaker ID.\\

\textbf{Stage 5: Annotations.}
    The majority of Persian TV shows lack subtitles, thus we use the commercial Aipaa's\footnote{\url{https://aipaa.ir/}} automatic speech recognition (ASR) to construct sentence-level approximated video transcription.\\
    
\textbf{Stage 6: Keyword selection and preprocessing.}
     Since our task is word-level we should select appropriate words for network training. Here are the precise steps of keyword selection and preprocessing:
    \begin{enumerate}
      \item \textbf{Words collection.}
      First we collect the words from each video's transcription and form the dictionary of vocabularies with their frequencies.
      \item \textbf{Keywords selection.}
      The vocabularies are sorted based on the frequencies in descending order. We select our keywords from the top frequent words.
      \item \textbf{Lemmatization.}
      In this step we recognize the group of words with the same root by the Persian lemmatization algorithm. After that, the word with the most frequency will be selected. The goal of this step is to increase the diversity of labels of the dataset.
      
      \item \textbf{Video split.}
      Finally, we split our videos based on the keywords that were chosen. Each video is padded with 200 milliseconds(ms), The first 100 milliseconds and the last 100 milliseconds of the video are included. 
    \end{enumerate}
     
\textbf{Stage 7: Face verification and Train/Test split.}
    To recognize faces, we employ ArcFace~\cite{deng2019arcface} and create face feature embeddings. The goal of this stage is to extract identities and make a speaker-independent dataset. Such datasets are useful for our task i.e. lip-Reading. We pick one of the programs as the test set and the rest of programs form the training set. This policy for the dataset split makes it roughly speaker-independent. To make speaker-independency complete, we compare the list of celebrities in the test set with the training set manually and remove the duplicated one in the test set.\

\begin{figure}[t]
  \centering
  \includegraphics[width=0.5\textwidth]{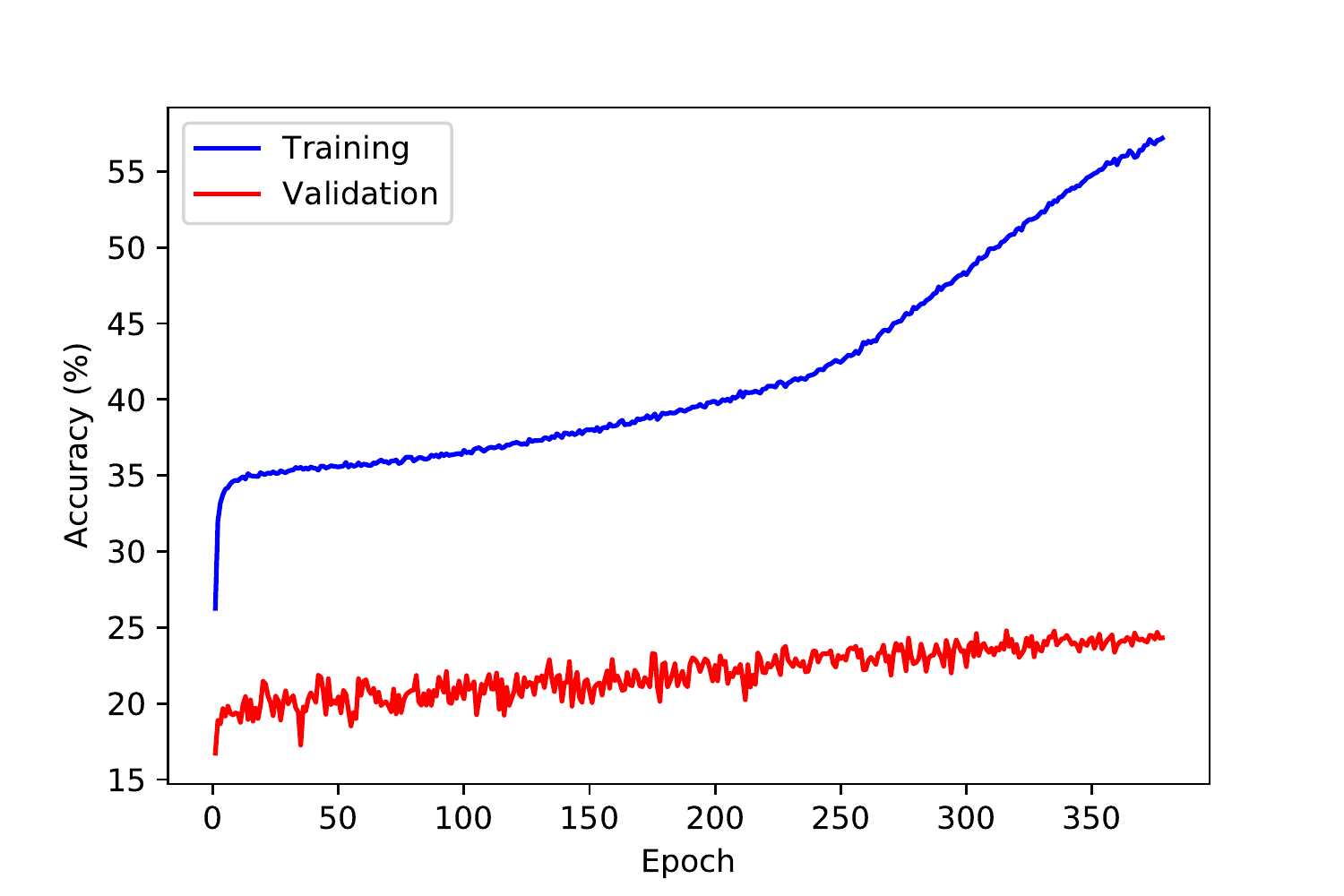}
  \vspace{-6mm}
  \caption{Training and Validation accuracies plot}
  \label{fig:av_hubert_acc_curves}
\end{figure}

\section{Experiments}
In this section, we describe baseline methods and show the performance of said methods on our dataset. We give details about training setups and the hardware that was used.

\subsection{Evaluation metrics}
We demonstrate the model performance by Top-k accuracy. In calculating this criteria, the model's prediction is correct if the actual class is among the k-classes with the highest score. In addition, we provide the macro average accuracy for each of the methods. The advantage of using the Macro f1-score is that it gives equal weight to all keywords, Regardless of the number of keyword occurrences in the test set.

\subsection{Evaluation on MS-TCN}
For the first experiment, we used the MS-TCN method. This method has obtained state-of-the-art results on the LRW dataset. For training, we used the official code provided in the GitHub repository.\footnote{\url{https://github.com/mpc001/Lipreading\_using\_Temporal\_Convolutional\_Networks}}

Table \ref{tab:acc} shows the results of this experiment in the first row. With micro and macro average criteria, the performance of this model on our data was \%13.46 and \%10.27, respectively.

\begin{figure}[t]
  \centering
  \includegraphics[width=0.5\textwidth]{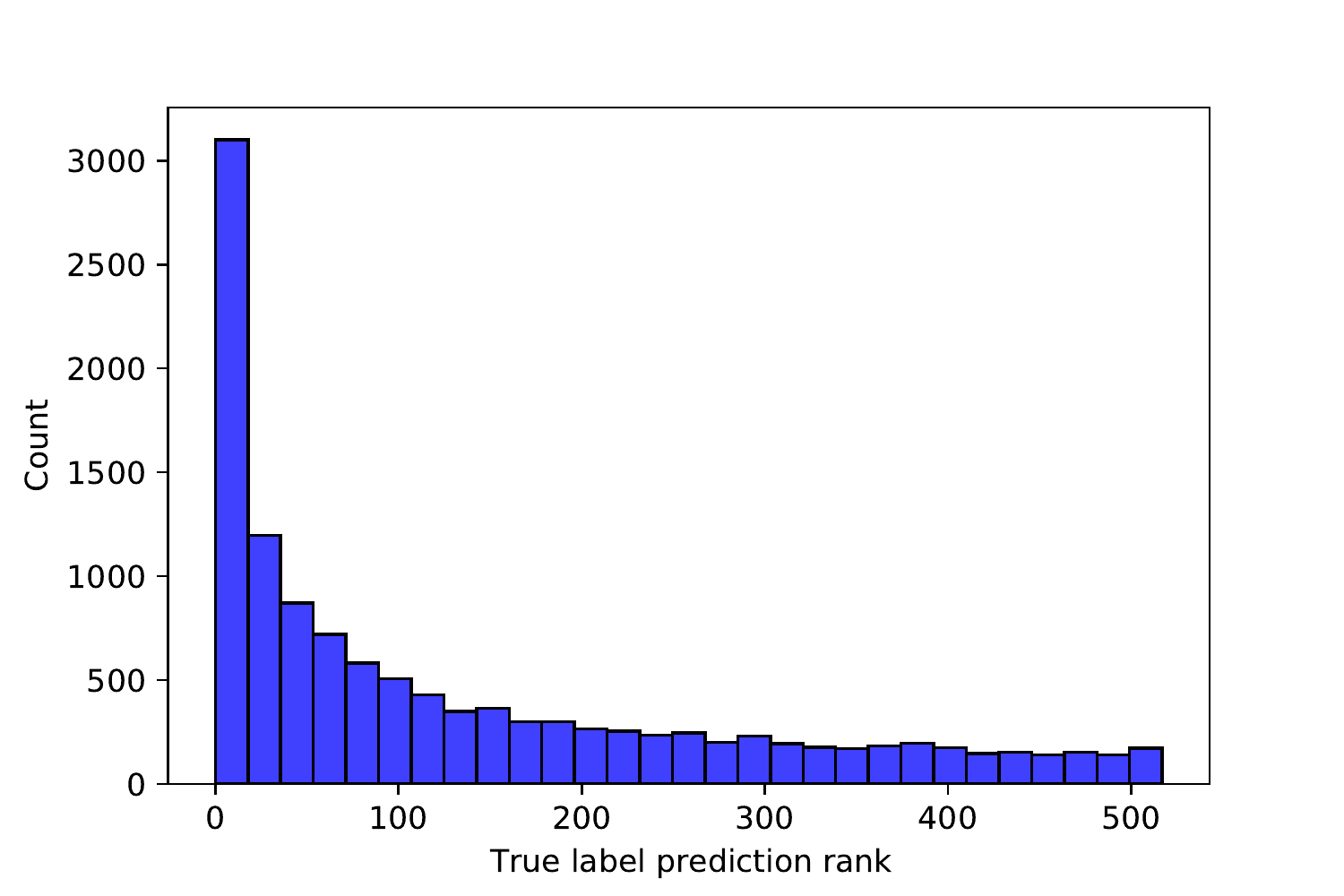}
  \vspace{-6mm}
  \caption{Histogram of true class prediction ranks.}
  \label{fig:rank_histogram}
\end{figure}

\subsection{Evaluation on AV-Hubert}
We utilized the AV-HuBERT approach to create another baseline for our dataset. This method is for lip-reading at the sentence level. In this case, we used it as a feature extractor.

Since word-level lip-reading is a classification task, the model must assign each input to a class. To this end, we obtain the embedding vector of each frame by feeding the input video to a pre-trained AV-HuBERT Base model, and these features are given to two linear layers. The average of logits values for each class were used to obtain probabilities for each class.

Figure \ref{fig:rank_histogram} shows the histogram of the true class prediction rank. We take each sample's actual class rank in the model's prediction into consideration to generate this histogram. For example, the value of this function for x=1 is 985, indicating that the real class is predicted as the second-highest score for this many samples in the test set. Based on the results in Table \ref{tab:acc} and Table \ref{tab:topk}, this model outperforms the previous method. Interestingly, the feature extraction model is trained only on English datasets. Table \ref{tab:The most frequent errors} shows the most common errors.

\textbf{Hyperparameters: }We used the default hyperparameters that the original code declared for all of our models. Also, we trained our models on NVIDIA GTX 1080 GPU with 8 GB memory capacity. The implementation of the code is based on PyTorch.

\begin{table}[t]
 \caption{Recognition results on PLRW}
  \centering
  \begin{tabular}{c | c | c}
    \toprule
    \toprule
    Model     & Micro Average Acc. & Macro Average Acc. \\
    \midrule
    MS-TCN & \%13.46 & \%10.27  \\
    AV-Hubert & \textbf{\%24.79} & \textbf{\%21.43}  \\
    \bottomrule
    \bottomrule
  \end{tabular}
  \label{tab:acc}
\end{table}

\begin{table}[t]
 \caption{Top-k accuracy on PLRW}
  \centering
  \begin{tabular}{c | c | c | c}
    \toprule
    \toprule
    Model      & Top-3 Acc. & Top-5 Acc. & Top-10 Acc.\\
    \midrule
    MS-TCN  & \%22.39 & \%28.53 & \%37.12 \\
    AV-Hubert  & \textbf{\%38.79} & \textbf{\%45.47} & \textbf{\%55.03} \\
    \bottomrule
    \bottomrule
  \end{tabular}
  \label{tab:topk}
\end{table}

\section{Conclusion}
In this paper, we presented a dataset for Persian word-level lip-reading. This large dataset is suitable for the proposed methods based on deep learning. We also tested the dataset with the best model of word-level lip-reading on the LRW benchmark. We demonstrated a novel method in which we achieved better accuracy through feature extraction by the AV-HuBERT model.

\section{Acknowledgement}

This project was mainly supported by Arman Rayan Sharif company, an AI company in Iran. 

\begin{table}[t]
 \caption{The most frequent errors}
  \centering
  \begin{tabular}{c | c | c}
    \toprule
    \toprule
    Number of Occurrences     & Ground Truth     & Prediction\\
    \midrule
    20 & \RL{برای}&  \RL{بعد}   \\
    17 & \RL{اولین}&  \RL{ولی}    \\
    15 & \RL{قبل}&  \RL{همه}    \\
    15 & \RL{خود}&  \RL{بود}    \\
    14 & \RL{همون}&  \RL{بود}    \\
    13 & \RL{بوده}&  \RL{بود}    \\
    13 & \RL{همین}&  \RL{همه}    \\
    12 & \RL{شما}&  \RL{بود}    \\
    12 & \RL{نظر}&  \RL{بعد}    \\
    12 & \RL{مربی}& \RL{همه} \\
    \bottomrule
    \bottomrule
  \end{tabular}
  \label{tab:The most frequent errors}
\end{table}

\bibliographystyle{IEEEtran}

\bibliography{mybib}

\end{document}